\documentclass[10pt, a4paper]{article}
\usepackage{lrec2022} 
\usepackage{multibib}
\newcites{languageresource}{Language Resources}
\usepackage{graphicx}
\usepackage{tabularx}
\usepackage{soul}
\usepackage{titlesec}
\titleformat{\section}{\normalfont\large\bfseries\center}{\thesection.}{1em}{}
\titleformat{\subsection}{\normalfont\SmallTitleFont\bfseries\raggedright}{\thesubsection.}{1em}{}
\titleformat{\subsubsection}{\normalfont\normalsize\bfseries\raggedright}{\thesubsubsection.}{1em}{}
\renewcommand\thesection{\arabic{section}}
\renewcommand\thesubsection{\thesection.\arabic{subsection}}
\renewcommand\thesubsubsection{\thesubsection.\arabic{subsubsection}}

\usepackage{epstopdf}
\usepackage[utf8]{inputenc}

\usepackage{hyperref}
\usepackage{xstring}

\usepackage{color}

\usepackage{times}
\usepackage{latexsym}
\usepackage{amssymb}
\usepackage{pifont}
\usepackage{graphicx}
\usepackage{verbatim}
\usepackage{xcolor}
\usepackage{array}
\usepackage{makecell}
\usepackage{booktabs}
\usepackage{subcaption}
\usepackage{microtype}
\usepackage{multirow}
\usepackage{hyperref}

\title{ELF22: A Context-based Counter Trolling Dataset to Combat Internet Trolls}
\name{
Huije Lee$^*$, 
Young Ju NA$^*{^\dagger}$, 
Hoyun Song, 
Jisu Shin, 
Jong C. Park
}

\address{
KAIST\\
Daehak-ro 291, Yuseong-gu, Daejeon, Republic of Korea\\
\{jae4258,hysong,jsshin,park\}@nlp.kaist.ac.kr\\
youngju073092@gmail.com\\
}

\abstract{
Online trolls increase social costs and cause psychological damage to individuals. With the proliferation of automated accounts making use of bots for trolling, it is difficult for targeted individual users to handle the situation both quantitatively and qualitatively. To address this issue, we focus on automating the method to counter trolls, as counter responses to combat trolls encourage community users to maintain ongoing discussion without compromising freedom of expression. For this purpose, we propose a novel dataset for automatic counter response generation. In particular, we constructed a pair-wise dataset that includes troll comments and counter responses with labeled response strategies, which enables models fine-tuned on our dataset to generate responses by varying counter responses according to the specified strategy. We conducted three tasks to assess the effectiveness of our dataset and evaluated the results through both automatic and human evaluation. In human evaluation, we demonstrate that the model fine-tuned on our dataset shows a significantly improved performance in strategy-controlled sentence generation.
 \\ \newline  }

\begin{document}

\maketitleabstract

\section{Introduction}

\def\thefootnote{*}\footnotetext{equal contribution}\def\thefootnote{\arabic{footnote}}

\def\thefootnote{$\dagger$}\footnotetext{Present Address: Young Ju NA, Université Sorbonne Nouvelle, 17 rue de Santeuil 75005 Paris}\def\thefootnote{\arabic{footnote}}

The International Telecommunication Union (ITU) stated that the number of internet users increased from 4.1 billion in 2019 to 4.9 billion in 2021, as 782 million people started to participate in cyberspace activities during the COVID-19 pandemic. The Pew Research Center in 2021 reported that 93\% of American adults participate in internet communities to interact with others. Together with the positive side of online communities that promote open discussion and facilitate social movements, the number of users experiencing online trolling or abuse has also increased. The Australia Institute \shortcite{australia2019trolls} reported the estimated social cost of online trolling and abuse at \$3.7 billion in 2019.

Online trolling is a malicious attempt to provoke strong reactions from an interlocutor. Trolls, it would seem, do not care what they are saying as long as they have an impact on the chosen target \cite{Fichman2016OnlineTA,mihaylov-nakov-2016-hunting,golfpapez17}. Regarding the issue of online trolls, there have been attempts by community managers such as Twitter, Facebook, and Reddit to delete troll comments or by users to ignore them \cite{sibai2015social}. However, compared to the troll attackers, the voice of the victim group is overall weak. For example, Asian hate speech related to COVID-19 showed a large difference in the number of tweets, as there were 3 times more users displaying hate speech than those displaying counter speech, and hateful bots were more active and more successful in attracting followers than counter-hate bots \cite{ziems2020racism}.

To counteract the problem posed by online trolling, we propose to adopt a method of actively countering trolls so as to regulate abusive language online. This allows the online discussion to keep going, rather than to see certain abusive messages deleted and the flow of conversation interrupted. As argued also by Golf-Papez and Veer \shortcite{golfpapez17}, counter responses help support the users and protect them from trolls by neutralizing the negative impact of trolling. In countering trolls, we also propose to follow the line of making more utterances against trolls \cite{richards2000counterspeech}, with controlled strategies \cite{Hardaker2015IRT}. 

Compared to counter speech studies \cite{qian-etal-2019-benchmark,chung-etal-2019-conan,tekiroglu-etal-2020-generating,laugier-etal-2021-civil}, research on automatic sentence generation to counter trolls is still in its infancy. Despite the abundance of troll detection datasets \cite{mihaylov-nakov-2016-hunting,wulczyn2017ex,atanasov-etal-2019-predicting}, there are few datasets with the rich contextual information needed to improve the quality of counter responses against trolls. Addressing this situation, we present a novel dataset, which can be used for automatically generating responses to combat trolls, acting as an “Elf bot”. For this purpose, we first collected pairs of troll comments and their counter responses in the Reddit community. We then formulated a scheme according to the seven counter response strategies \cite{Hardaker2015IRT}: \textit{engage}, \textit{ignore}, \textit{expose}, \textit{challenge}, \textit{critique}, \textit{mock}, and \textit{reciprocate}. In order to help annotators understand counter response strategies more clearly, we categorized trolls into two types \cite{hardaker2013uh}: overt and covert trolls. We then examined the effectiveness of our annotated dataset. Finally, we assessed the quality of counter responses generated by strong baselines for a given response strategy through automatic and human evaluation. 

We highlight three major contributions of our paper: 1) We provide the first dataset for countering trolls labeled with counter strategies and rich contextual text; 2) we show the use case of the dataset with robust baselines for three tasks: binary classification for troll strategies, multi-class classification for counter response strategies, and counter response generation; and 3) we demonstrate the effectiveness of the dataset through automatic and human evaluation on counter responses generated by models fine-tuned over this dataset.

The rest of the paper is organized as follows: In Section~\ref{sec:related_work}, we introduce related work on troll definition, troll detection, and counter speech. We explain the detailed labels of troll comments and counter responses for data annotation in Section~\ref{sec:data_annotation}. We describe the process of collecting and labeling for the dataset with statistics in Section~\ref{sec:dataset}. In Section~\ref{sec:experiment}, we describe three models trained over our dataset with their performances and demonstrate the effectiveness of our dataset through automatic and human evaluation. Finally, we show conclusion and future work in Section~\ref{sec:conclusions}.

\section{\label{sec:related_work}Related Work}

Trolls can be criminalized and prosecuted for unlawful behavior and illegal activities that include intentional provocation of anguish \cite{golfpapez17}. Trolling behaviors encompass the intention that causes emotional distress to the targets\footnote{The 2021 Florida Statutes 784.048, retrieved from http://www.leg.state.fl.us/Statutes/index.cfm}, and are grossly offensive, false, indecent, menacing, provocative or disturbing\footnote{Communications Act 2003, retrieved from https://www.legislation.gov.uk/ukpga/2003/21/section/127} that can amass to a criminal offense. Trolls communicate to manipulate people’s opinions and make mischievous attempts to elicit reactions from their targets, or, interlocutors. Trolls, or attention-seekers, aim to provoke responses from their targets through a variety of communication tactics such as hyper-critism, shaming, sarcasm, blackmailing, and publishing of private information \cite{Fichman2016OnlineTA,mihaylov-nakov-2016-hunting,golfpapez17}.

Taking into consideration these diverse tactical aspects of trolls, several studies constructed datasets with labeled troll types in order to detect trolls in online communities \cite{mihaylov-nakov-2016-hunting,wulczyn2017ex,atanasov-etal-2019-predicting}. Other studies have proposed a troll detection model by using an annotated dataset of troll comments \cite{kumar14,georgakopoulos2018convolutional,chun19,atanasov-etal-2019-predicting}. Jigsaw and Google ran the Perspective project, providing APIs\footnote{https://www.perspectiveapi.com} to score the text’s toxicity, together with a troll dataset\footnote{https://www.kaggle.com/c/jigsaw-toxic-comment-classification-challenge}. While these studies focused only on the detection of trolls, we categorized trolls by their overtness and identified the best counter strategies by their types.

Other than such numerical values of toxicity of contents, there are some studies that focused on counter speeches to combat abusive language online. Mathew et al. \shortcite{mathew2018analyzing,mathew2019thou} and Chung et al. \shortcite{chung-etal-2019-conan} proposed datasets for counter speech with a taxonomy proposed by Benesch et al. \shortcite{benesch16}. Subsequent studies have shown that these datasets have significantly improved the performance of counter speech generation models \cite{chung2020italian,tekiroglu-etal-2020-generating,laugier-etal-2021-civil}. Unlike previous studies, however, we focused on trolling to identify counter strategies against each troll type for diversified counter response generation.

\section{\label{sec:data_annotation}Data Annotation Scheme}

\begin{table}[t!]
\begin{tabular}{|l|l|}
\hline
\multicolumn{1}{|c|}{} & \multicolumn{1}{c|}{\makecell[l]{
\textbf{Aggress} is directly cursing or \\ swearing others without any \\
justification
}} \\ \cline{2-2} 
\multicolumn{1}{|c|}{\multirow{2}{*}{\makecell[c]{Overt \\ strategies}}}                       & \makecell[l]{
\textbf{Shock} is throwing an ill-disposed \\
or prohibited topic that is avoided for \\
political or religious reasons.
}     \\ \cline{2-2} 
\multicolumn{1}{|c|}{}                       & \makecell[l]{
\textbf{Endanger} is providing \\
disinformation with the intent to harm \\
others, and discovering this purpose by \\
others.
}                      \\ \hline
\multirow{3}{*}{}                      & \makecell[l]{
\textbf{Antipathize} is creating a \\
sensitive discussion that evokes an \\
emotional and proactive response in \\
others.
}                      \\ \cline{2-2} 
\makecell[c]{Covert \\ strategies}           & \makecell[l]{
\textbf{Hypocriticize} is excessively \\
expressing disapproval of others or \\
pointing out faults to the extent that \\
it feels intimidating to others.
}                      \\ \cline{2-2} 
                                             & \makecell[l]{
\textbf{Digress} is making a discussion \\
to be derailed into irrelevant or toxic \\
subjects.
}                      \\ \hline
\end{tabular}
\caption{\label{tab:trolltypes} Types of troll behaviors \protect\cite{hardaker2013uh}}
\vspace{-0.1in}
\end{table}

\begin{table*}[t!]
\centering
\begin{tabular}{l|l|l|l|ll}
\hline
\multicolumn{1}{c|}{\bf Title} & 
\multicolumn{1}{c|}{\bf Post} & 
\multicolumn{1}{c|}{\bf Troll comment} & 
\multicolumn{1}{c|}{\bf Response} & 
\multicolumn{1}{c}{\bf TL} & 
\multicolumn{1}{c}{\bf RL} 
\\ \hline
\makecell[l]{
If I'm not \\
going to \\
vaccinate \\
myself, why?}    &
\makecell[l]{
Just heard that NC is\\
considering giving \\
portions of doses \\
on-hand back to feds. \\
If you've decided to \\
not get jabbed, what's \\
your reasoning?
}   & 
\makecell[l]{
I am glad you and a \\
bunch of dumbs live \\
in a nation that \\
lauds your ignorance. \\
covid is going to \\
kill some of you \\
idiots moving \\
forward.\\
} & 
\makecell[l]{
I got my shots, TYVM. \\
I asked in general to \\
attempt an antagonizing \\
dialog with folks. Please \\
try better, and remember \\
you catch far more flies \\
with honey than vinegar.
}       
& 1      & 5     \\ \hline
\makecell[l]{
I think you \\
guys complain\\
too much
}    & 
\makecell[l]{
Everday I see posts \\
like ``there's too much \\
damage", ``too much \\
mobility", ... I don't \\
know. LoL has 140 \\
champions and they all \\
sit between 45-55\% \\
winrate, Riot got\\
the one of the most\\
popular games out\\
there for 10+ years.
}   & 
\makecell[l]{
cringe post you can \\
still delete this
} & 
\makecell[l]{
Cringe comment You \\
can still delete this
}       
& 1      & 7      \\ \hline
\end{tabular}
\caption{Examples of collected Reddit posts, along with annotated strategies. TL: Troll strategy label; RL: Response strategy label. The number 1 of the TL column indicates overt troll. The numbers 5 and 7 of the RL column indicate \textit{critique} and \textit{reciprocate}, respectively.}
\label{tab:postexamples}
\end{table*}

\subsection{Troll Strategy}
A troll teases people to make them angry, offends people, wants to dominate any single discussion, or tries to manipulate people’s opinions \cite{mihaylov-nakov-2016-hunting}. Trolling shows a specific type of aggressive online behavior involving antagonism for the sake of amusement such as a deliberate, deceptive and mischievous attempt to provoke a reaction from other online users \cite{articleHerring2002,Shin2008MoralityAI,articleBinns2012,hardaker2013uh,golfpapez17}.
In particular, Hardaker \shortcite{hardaker2013uh} introduced the notion of overt and covert trolls with a detailed scale of trolling bahaviour in six categories: \textit{aggress}, \textit{shock}, \textit{endanger}, \textit{antipathize}, \textit{hypocriticize}, and \textit{digress}. Since we focus on counter responses to combat trolls, we used two super-categories, overt and covert strategies, to classify trolls by their detailed tendencies (see Table~\ref{tab:trolltypes}).

\subsection{Response Strategy}
The function of countering trolls is to be a “virtually capable guardian”, who can neutralize the impact of trolling on targets. Note that when devising a counter response, it is not always possible to identify who triggered the action of trolling through comments \cite{hardaker2013uh}. Therefore, we followed the taxonomy of using seven response strategies for counter responses, introduced by Hardaker \shortcite{Hardaker2015IRT},  as follows.

\textbf{Engage} strategy indicates acting in accordance with the troll's intentions. This strategy involves accepting the troll's opinions positively and responding sincerely to the troll. From the third party's point of view, however, the users employing this strategy could be considered to have fallen prey to the troll.

\textbf{Ignore} strategy has the goal of not giving what the troll wants. Users attempt to warn others about doubtful behaviors of the troll. The strategy is to deter trolls from their malicious intention so that the trolls leave the discussion. For this strategy to work, it is necessary to be proactive in discovering the troll's intentions.

\textbf{Expose} strategy takes a stance either going against the troll's opinion or doubting the authenticity of the troll’s information. This strategy has the risk of being trolled by responding to the message of trolls; however, it could keep the discussion consistent and healthy.

\textbf{Challenge} strategy provides direct opposition to the troll with an aggressive response. Users often use emotional language to express their hostility. This strategy has a similar purpose to \textit{expose} in terms of confronting trolls, but \textit{challenge} tries go more strongly against what trolls have said.

\textbf{Critique} strategy goes beyond discovering the true intention of the troll, and overtly judges the troll's behavior as degrading and uncreative. This strategy weakens the troll's effectiveness by evaluating its actions rather than the content of the troll comment.

\textbf{Mock} strategy involves making the troll scorned and ridiculed. With mocking, users attempt to isolate the troll while strengthening the cohesion of discussion participants.

\textbf{Reciprocate} strategy attempts to attack the troll by pretending to be another troll. This strategy takes a more radical stance than the \textit{challenge} strategy in that the purpose is to attack the troll rather than to self-defend against the troll.

\section{\label{sec:dataset}ELF22 Dataset}

We present a troll-response pairwise dataset, ELF22, along with contextual text from posts in the Reddit community. Our dataset contains posts with troll comments and counter responses. 
Each troll comment and counter response pair is annotated with a proposed strategy according to the data annotation scheme (see Section~\ref{sec:data_annotation}). In this section, we present a detailed description of our data labels, the data collection process, data statistics, and ethical considerations of the given dataset.

\subsection{Data Collection}

For the data collection, we crawled the posts that include troll comments from Reddit using Pushshift API \cite{baumgartner2020pushshift}. We describe how we limited our conditions as well as some of the significant parts of our data collection process. First, we extracted our data from diverse subreddit communities, taking into account the fact that trolls are “light speakers”, who are not, or do not have the capacity to be, deeply involved in a conversation, as opposed to “heavy speakers”, who present logical arguments to discuss with others \cite{choi15}. Second, we focused on the importance of the fast interactions and interchange of messages amongst online community users, which aggravates the propagation of troll comments, since virality, responsiveness and volume are the three important factors in terms of conversation characteristics of Reddit \cite{choi15}. 
Third, we considered the title and body text of the post as context. In order to exclude the possibility of disconnection in the context history, we only extracted root comments if they were described to be by trolls. 

To collect unified data types, we limited our range of extracting data of online trolls. Concretely, we set the range of thumbs down of troll comments from -2 to -15. 
The upper limit of a downvoted comment was set to -15, as it excludes comments heavily biased against community tendencies. Next, we extracted the highest-scored comment of thumbs up among the following comments from the root comment as a counter response reference.

We discarded posts if they contained URLs, included pictures, had more than 512 characters, or used language other than English. We then selected the posts whose number of characters is at least 12. Samples of the collected data are shown in Table~\ref{tab:postexamples}.

\subsection{Data Annotation}
For this annotation task, we recruited 12 annotators in total who are familiar with the Reddit website. Native English speakers or those whose education took place in a country where English is used as an official language participated in this experiment since our collected dataset is in English.

The annotation phase of the dataset is organized in two sessions, each of which took 7 days to complete. We asked all the annotators to follow the guidelines which include the definition of trolls, the importance of combating them, and an explanation of each strategy with examples (see Section~\ref{sec:data_annotation}). Since these sessions were conducted online, we devised a strictly controlled environment in order to obtain a qualified dataset. Each annotator had his or her own personalized Google sheet link to annotate a scheduled amount of the dataset every day. Each day, annotators submitted their labeled dataset in order to receive a confirmation from our team. In addition to this, we regularly opened online QA sessions to help annotators, at the same time ensuring that they were faithfully engaged and fully familiar with the task.

For the first annotation session, annotators were given Reddit posts with subreddit names, titles, and body text, including two target sentences to annotate: a ‘troll comment’ and its ‘response’. Annotators were asked to read the troll comment and label it according to whether they considered it to be a ‘covert’ or ‘overt’ type of troll. They were then asked to read the response, which was considered to be a combating sentence against the troll, before giving it a label according to the 7 strategies from the guidelines. We gave the annotators an option to propose their own choice of a counter response strategy against the troll if they believed that the given ‘response’ was considered ineffective. Since the entries of the dataset were systematically distributed so that three annotators could work on the same entries of the dataset, we found disagreements among the annotators after the first session. Therefore, for the second annotation session, if all the three annotators gave a different label to the text, we asked them to re-consider their choices by discussing among them and re-labeling the data. We chose the majority labels otherwise.

Table~\ref{tab:postexamples} shows examples of the five elements in each entry: subreddit (omitted in Table~\ref{tab:postexamples}), title, post, troll (troll comment), and response (counter response). The first entry is a post about vaccination and asking the opinion of others regarding the decision of refusing to be vaccinated. The comment is identified as a troll by the subreddit users, which we interpret as a troll that indirectly aggresses other locutors but does not directly name any reddit user.

\subsection{Data Statistics}

\begin{table}[t!]
\centering
\begin{tabular}{ccc}
\toprule
 & \bf Avg. \# words & \bf Max. \# words \\
\midrule
\multicolumn{1}{l}{Title}  & \multicolumn{1}{r}{12.2}  & \multicolumn{1}{r}{82} \\
\multicolumn{1}{l}{Body}  & \multicolumn{1}{r}{57}  & \multicolumn{1}{r}{205} \\
\multicolumn{1}{l}{Troll comment}  & \multicolumn{1}{r}{38.4}  & \multicolumn{1}{r}{126} \\
\multicolumn{1}{l}{Response}  & \multicolumn{1}{r}{25.6}  & \multicolumn{1}{r}{128} \\
\midrule
\multicolumn{1}{l}{Total}        & \multicolumn{1}{r}{133.2}  & \multicolumn{1}{r}{-} \\ 
\bottomrule
\end{tabular}
\caption{The average number of words in each element of posts}
\label{tab:average_words}
\end{table}

\begin{table}[t!]
\centering
\begin{tabular}{ccccc}
\toprule
\bf Strategy  & \bf Train & \bf Validation & \bf Test & \bf Total \\
\midrule
\multicolumn{1}{l}{Overt}  & \multicolumn{1}{r}{2,331}  & \multicolumn{1}{r}{166}  & \multicolumn{1}{r}{340}  & \multicolumn{1}{r}{2,837} \\
\multicolumn{1}{l}{Covert}  & \multicolumn{1}{r}{3,020}  & \multicolumn{1}{r}{407}  & \multicolumn{1}{r}{422}  & \multicolumn{1}{r}{3,849} \\
\midrule
\multicolumn{1}{l}{Engage}  & \multicolumn{1}{r}{2,134}  & \multicolumn{1}{r}{340}  & \multicolumn{1}{r}{343}  & \multicolumn{1}{r}{2,817} \\
\multicolumn{1}{l}{Ignore}  & \multicolumn{1}{r}{216}  & \multicolumn{1}{r}{10}  & \multicolumn{1}{r}{35}  & \multicolumn{1}{r}{261} \\
\multicolumn{1}{l}{Expose}  & \multicolumn{1}{r}{1,131}  & \multicolumn{1}{r}{75}  & \multicolumn{1}{r}{153}  & \multicolumn{1}{r}{1,359} \\
\multicolumn{1}{l}{Challenge}  & \multicolumn{1}{r}{818}  & \multicolumn{1}{r}{48}  & \multicolumn{1}{r}{71}  & \multicolumn{1}{r}{937} \\
\multicolumn{1}{l}{Critique}  & \multicolumn{1}{r}{340}  & \multicolumn{1}{r}{31}  & \multicolumn{1}{r}{52}  & \multicolumn{1}{r}{423} \\
\multicolumn{1}{l}{Mock}  & \multicolumn{1}{r}{592}  & \multicolumn{1}{r}{51}  & \multicolumn{1}{r}{96}  & \multicolumn{1}{r}{739} \\
\multicolumn{1}{l}{Reciprocate}  & \multicolumn{1}{r}{120}  & \multicolumn{1}{r}{18}  & \multicolumn{1}{r}{12}  & \multicolumn{1}{r}{150} \\
\midrule
\multicolumn{1}{l}{Total}  & \multicolumn{1}{r}{5,351}  & \multicolumn{1}{r}{573}  & \multicolumn{1}{r}{762}  & \multicolumn{1}{r}{6,686} \\ 
\bottomrule
\end{tabular}
\caption{Statistics of the dataset according to troll strategies and response strategies, divided into train, validation, and test datasets.}
\label{tab:strategy_num}
\end{table}

First, we collected from Reddit 5,700 posts that include 2,198 unique subreddits. The top 10 subreddits of the collected posts were: \textit{unpopularopinion}, \textit{wallstreetbets}, \textit{teenagers}, \textit{nba}, \textit{relationship\_advice}, \textit{DestinyTheGame}, \textit{NoStupidQuestions}, \textit{Genshin\_Impact}, \textit{fo76}, and \textit{NonewNormal}. 

We removed posts with incomplete composition during the annotation process. If a sentence was labeled differently such as \textit{expose}, \textit{challenge}, and \textit{reciprocate} by three annotators even after the second annotation session, we have considered it as outliers and discarded it. We only extracted the labeled posts that obtained the majority agreement among three annotators.
After filtering, we finalized our dataset with a total 5,535 pairs of troll comments and counter responses, together with 1,151 ‘your response’ sentences that were made by the annotators.

\begin{figure}[!t]
\centering
\includegraphics[scale=0.177]{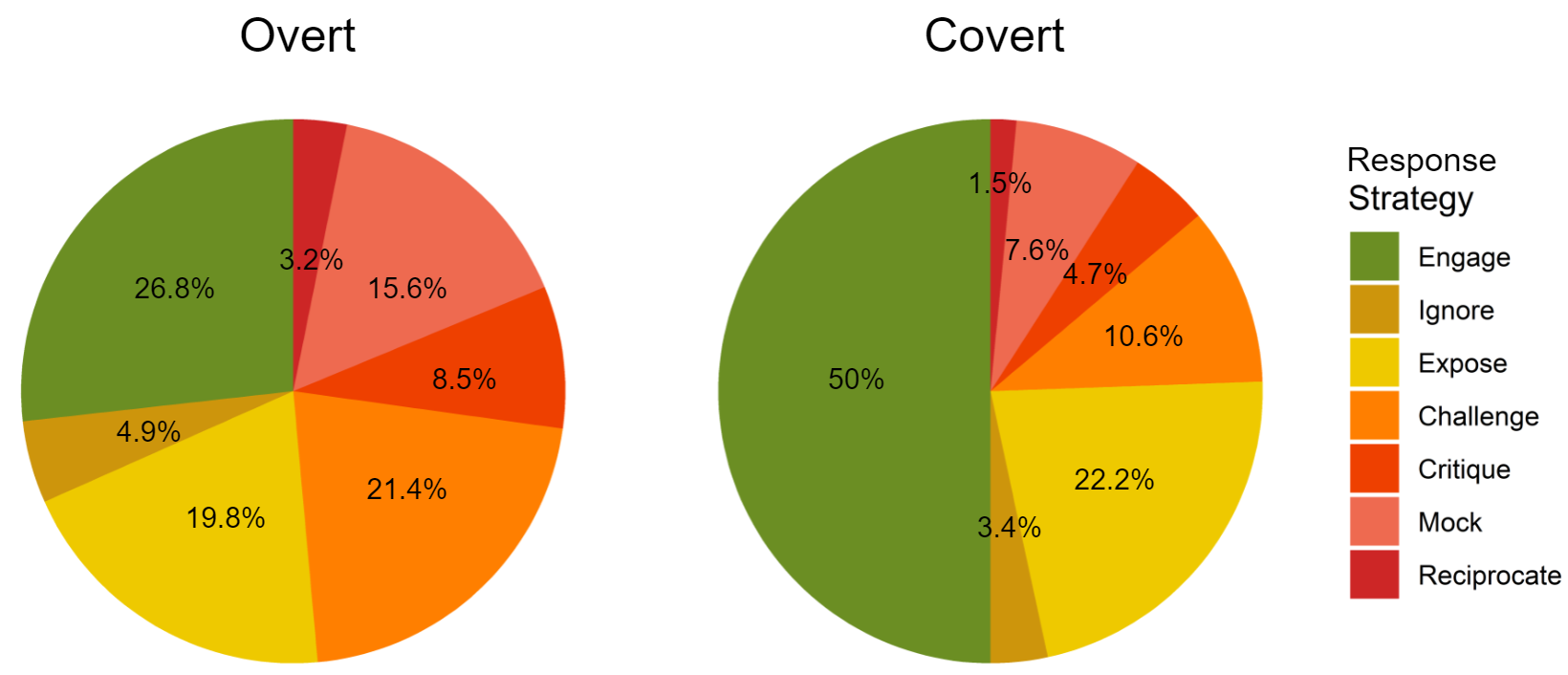}
\caption{Distribution of response strategies to overt troll (left) and covert troll (right)}
\label{fig:datastatistics_conditioned}
\end{figure}

Table~\ref{tab:average_words} shows information about the average and maximum numbers of words of our dataset, where the average numbers of words of contextual texts, troll comments, and responses are 69.2, 38.4, and 25.6, respectively.

Figure~\ref{fig:datastatistics_conditioned} shows the distribution of response strategy types in our dataset. We see that Reddit users often used the \textit{engage} strategy, followed by the \textit{exposure} and \textit{challenge} strategies. On the other hand, the \textit{ignore} and \textit{reciprocate} strategies occupied only a small portion of the dataset.
 
Table~\ref{tab:strategy_num} presents the statistics of our dataset with respect to each label. The labeled posts are randomly split into training (80\%), validation (10\%), and test (10\%). We note that each post has a unique post number in accordance with trolling comments. This means that we assigned the same post number for the comments of countering trolls and `your response' sentences. The consequent distribution of entries of the ELF22 dataset was training (80\%), validation (8.6\%), and test (11.4\%).

To check the validity and consistency of the collected dataset, we performed the Inter-Annotator Agreement (IAA). The Fleiss’s Kappa ($\kappa$) value of the troll type label of our dataset was 0.539, and the value of the counter strategy label was 0.465, achieving a ‘moderate’ agreement between the two labels \cite{mchugh2012interrater}.

\subsection{Ethical Consideration}

Our annotation task was approved by the Institutional Review Board (IRB)\footnote{Approval number: KH2021-126}. All participants of annotation tasks indicated their understanding of the procedure for the annotation and acknowledged their agreement to participate. We report that all data are anonymized, and we plan to ensure that future researchers follow ethical guidelines, with the necessary information collected as the occasion demands.

The goal of our work is to categorize responses against trolls in online conversations and support the development of generation bots for countering trolls according to seven strategies proposed in this paper.
As shown in Tables~\ref{tab:postexamples},~\ref{tab:morepostexamples}, and~\ref{tab:morepostexamples2}, our dataset may contain sarcastic and aggressive language. We tried to observe how they communicate as-is, even though it could include socially biased content or hate speech. We expect that our dataset will be helpful for future work to study effective methodologies for responding to troll language online.

When deploying language models for response generation on our dataset, the models could generate unintended results, such as social and racial biases. There could be solutions including list-based token filtering, prompt conditioning \cite{keskar2019ctrl}, domain-adaptive pre-training \cite{gururangan-etal-2020-dont}, or any detoxification techniques \cite{gehman20} to avoid unintended response generation in accordance with the research purpose.

\section{\label{sec:experiment}Experiments}

\begin{table*}[t!]
\centering
\begin{tabular}{l|cccc|cccc}
\hline
\multicolumn{1}{c|}{\multirow{1}{*}{}} & \multicolumn{4}{c|}{\bf Task A} & \multicolumn{4}{c}{\bf Task B}  \\
\cline{2-9} 
\multicolumn{1}{c|}{} & \bf P & \bf R & \bf wF1 & \bf MF1 & \bf P & \bf R & \bf wF1 & \bf MF1 \\
\hline
 SVM    & 0.58 & 0.58  & 0.58 & 0.57 & 0.32 & 0.34 & 0.33 & 0.19 \\
 RF     & 0.60 & 0.59  & 0.54 & 0.52 & 0.30 & 0.45 & 0.28 & 0.09 \\
 BERT   & \bf0.64 & 0.63  & 0.63 & 0.63 & \bf0.48 & \bf0.47 & \bf0.47 & 0.27 \\
 RoBERTa& 0.64 & \bf0.64  & \bf0.64 & \bf0.64 & 0.46 & 0.42 & 0.42 & \bf0.28 \\
\hline
\end{tabular}
\caption{Evaluation results of the baselines on two classification tasks A and B. The best scores in each metric are highlighted in \textbf{bold}. P: weighted precision; R: weighted recall; wF1: weighted F1 score; MF1: macro F1 score.}
\label{tab:result_ab}
\end{table*}

\begin{table}[t!]
\centering
\begin{tabular}{l|cccc}
\hline
\multicolumn{1}{c|}{\multirow{1}{*}{}} & \multicolumn{4}{c}{\bf Task C}  \\
\cline{2-5} 
\multicolumn{1}{c|}{} & \bf R-L & \bf B-1 & \bf M & \bf BS \\
\hline
 GPT-2          & 0.04 & 0.04  & 0.08 & 0.29 \\
 BART           & 0.08 & 0.08  & \bf0.16 & \bf0.41 \\
 DailoGPT-ELF22 & 0.06 & 0.14  & 0.04 & 0.35 \\
 GPT-2-ELF22    & 0.06 & 0.15  & 0.04 & 0.34 \\
 BART-ELF22     & \bf0.10 & \bf0.15 & 0.09 & 0.40 \\
\hline
\end{tabular}
\caption{Automatic evaluation results of the baselines on response generation task C. The best scores in each metric are highlighted in \textbf{bold}. R-L: Rouge-L F1 score; B-1: BLEU-1; M: METEOR; BS: BERTScore.}
\label{tab:result_c}
\end{table}

\subsection{Experimental Setup}
In this section, we describe three tasks that are used to exploit our dataset: troll strategy classification as binary classification, response strategy classification as multi-label classification, and generating counter responses as text generation.

\textbf{Task A. troll strategy classification} 
This task involves classifying troll comments with titles and body texts and stating whether the comments are overt trolls or covert trolls. We implemented support vector machine (SVM) and random forest (RF) as dictionary-based classifiers, and BERT \cite{devlin-etal-2019-bert} and RoBERTa \cite{liu2019roberta} as pre-trained transformer-based classifiers. We utilized the pre-trained transformer models from the Huggingface library \cite{wolf2019huggingface}. The hyperparameter setting for each model is as follows:
 
\textbf{SVM} We used a linear kennel with a hyperparameter $c$ value of 0.01 out of [0.01, 0.1, 1, 10, 100, 1000, 10000] and a gamma of ‘auto’. 

\textbf{RF} We used a hyperparameter $d$ value of 100 out of [0.01, 0.1, 1, 10, 100, 500, 1000].

\textbf{BERT} We used ‘bert-base-cased’ pre-trained checkpoint and fine-tuned the model on our train dataset with 10 epochs. The size of the batch was 32, the learning rate was $2e$-$05$, the weight decay was $1e$-$2$, the warm-up steps were 100, the drop-out rate was set to 0.1, and the max sequence length was 512, which are the same as those that Devlin et al. \shortcite{devlin-etal-2019-bert} used to fine-tune models.

\textbf{RoBERTa} We used a ‘roberta-base’ pre-trained checkpoint with a similar number of model parameters to BERT. We followed hyperparameters of the same values as BERT.

For dictionary-based classifiers, the words with the top-512 frequencies from the title, body text, and troll comment were included in a bag of words as input. 
We prepared a list of candidate hyperparameters for a grid search and selected the one with the best weighted F1 score in the validation dataset. Training times of BERT and RoBERTa were about 10 minutes on one NVIDIA A100-PCIE-40GB. We reported the average results under 5 fine-tuning runs from our test dataset. 

\textbf{Task B. response strategy classification} 
This task involves classifying responses with titles, body texts, and troll comments, stating whether the responses are used in one out of 7 counter strategies. We experimented with the same baselines used in Task A. Aside from 0.1 being selected for SVM $c$, we used the same hyperparameter settings for each model as in Task A.

\textbf{Task C. counter response generation} 
This task involves generating responses with titles, body texts, troll comments, and labeled strategies. We implemented the following competitive baseline models:
 
\textbf{BART} \cite{lewis-etal-2020-bart} is a transformer encoder-decoder model that is trained on Common Crawl News, Wikipedia, book corpus, and stories. We employed a ‘bart-base’ model that contained 6 encoder layers and 6 decoder layers. The size of the batch was 32.

\textbf{GPT-2} \cite{radford2019language} is a transformer decoder model that is trained on Common Crawl Webtext. We employed a ‘gpt2’ model that contained 12 decoder layers. The size of the batch was 16.

\textbf{DialoGPT} \cite{budzianowski-vulic-2019-hello} is an additionally pre-trained GPT-2 model on a dialog corpus. Fine-tuning configuration is the same as that of the GPT-2 model.

We fine-tuned the models over 20 epochs with a learning rate of $2e$-$04$, a weight decay of $1e$-$2$, a drop-out rate of 0.1, a max input sequence length of 768, and a max output response length of 128.

In the decoding phase, we employed a beam search with the number of beams set to 5 and with blocking trigram repeats, which is widely used to avoid low-quality generation results \cite{klein-etal-2017-opennmt,paulus2017deep}. Based on the GPT2 tokenizer, the average token lengths of train, validation, and test input were 124.2, 122.9, 121, respectively. The average token lengths of train, validation, and test ground-truth responses were 28.1, 29.1, 25.5, respectively. Training times of GPT2, DialoGPT, and BART were each about 30 minutes on one NVIDIA A100-PCIE-40GB.

\subsection{Automatic Evaluation}

\textbf{Tasks A and B.} For the classification tasks, we evaluated the weighted Precision (P), Recall (R), F1 score (wF1), and macro F1 score (MF1).

\textbf{Task C.} For the generation task, we performed automatic evaluation and human evaluation. On Task C., we used ROUGE-L F1 score \cite{lin-2004-rouge}, BLEU-1 \cite{papineni2002bleu}, METEOR \cite{banerjee2005meteor}, and BERTScore \cite{bert-score} for the automatic evaluation. ROUGE-L F1, BLEU-1, and METEOR are based on n-gram overlaps that evaluate the similarity between the ground-truth response and the predicted one. BERTscore is a machine learning based automatic metric that captures semantic similarities by comparing their contextual embeddings.

\subsection{Results}

Table~\ref{tab:result_ab} shows the performance of the models on Tasks A and B. The models are fine-tuned on our train dataset and evaluated against the ground-truth labels for 762 test samples. The pre-trained transformer models BERT and RoBERTa outperformed the machine learning-based models SVM and RF. It implies that labeled strategies in our dataset are independent of simple lexical features, while sufficiently considering contextual text. For Task A, the wF1 scores of SVM and RF were close to 0.5, slightly higher than the random selection performance of binary classification. For Task B, we observed that the transformer-based models had about 50\% higher wF1 performance than the machine learning-based models. We speculate that Task B needs to consider semantic features to improve the performance. We also see that SVM and RF show low F1 performance on \textit{reciprocate} and \textit{ignore} strategies that contain a small number of train samples, resulting in poor MF1 performance. On the other hand, BERT and RoBERTa show robust performance with the unbalanced dataset.

Table~\ref{tab:result_c} shows the performance of the generation models on Task C. To confirm the effectiveness of the proposed dataset, we experimented with two pre-trained models and three fine-tuned models. Three fine-tuned models showed higher BLEU scores than pre-trained models. We find that BART-ELF22 shows the highest ROUGE and BLEU scores, indicating that words in generated response have the most overlap with ground-truth references. BART showed the highest METEOR performance, which means that it achieved high n-gram precision and recall with the ground-truth references. Aside from BLEU, BART and BART-ELF22 showed high performance on ROUGE, METEOR, and BERTScore. The high BERTScores of BART and BART-ELF22 imply that the generated sentences of the two models derive outputs semantically similar to the references.

\subsection{Human Evaluation}

To verify that the proposed dataset is enough to train a generation model which produces counter sentences against trolls with a specified response strategy, we conducted human evaluation on counter response generation models. Automatic evaluation metrics are related to the quality of generated texts, but they underestimate texts with a smaller vocabulary and heavily rely on ground-truth references. In particular, this phenomenon is frequently observed due to the characteristics of the Reddit community, which has a lot of free discussions and dialogues. To demonstrate the effectiveness of our dataset while embracing this phenomenon, we evaluated the counter response texts based on two criteria: relevance and compatibility.

First, we experimented with how well our dataset could be used to train a generation model to generate a counter response of a certain strategy when there is a corresponding golden response strategy and text. We evaluated generated texts from the `relevance' point of view \cite{zhu-bhat-2021-generate}. Relevance includes an assessment of the following two factors: 1) how coherent or contextually matched the generated counter response is with the troll text, and 2) how well the generated text counters troll text.  We ranked differently generated responses to assess relevance. The higher rank refers to more contextually coherent and effective responses to troll comments. On the other hand, the lower rank refers to responses that may be out of context and endanger the troll victims.

Second, we assessed how effectively our dataset could help train a generation model to create counter responses that correspond to strategies not in the dataset from the `compatibility' perspective. The dataset consists of pairs of a troll comment and its response, with only one golden response strategy and response text to one troll comment. In this regard, we checked if it was necessary to evaluate whether the dataset would be sufficiently effective for the model to generate responses corresponding to various but untrained response strategies against a troll text. Compatibility captures how well the model generates responses to given strategies, scored on a Likert scale of 1 to 5. If a counter response is related to the given strategy and close to a human response, the response gets a higher score.

A total of 5 evaluators fluent in English and familiar with Reddit forums participated in this human evaluation task. These five evaluators were equally assigned with 100 randomly selected samples from our dataset. Before the evaluation process, all evaluators were introduced to the definition of trolls and the seven strategies of countering trolls.

All the five sentences in a single post were evaluated by five evaluators: three response sentences for relevance evaluation and two response sentences for compatibility evaluation. The three response sentences for relevance evaluation include: (1) the ground-truth response; (2) the GPT-2 generated sentence; and (3) the BART-ELF22 generated sentence\footnote{We also performed a test on BART, but the result is excluded from the evaluation because the output is easily discerned. We employed the BART-ELF22 model that performed better in the automatic evaluation of Task C.}. Two response sentences include the generation of GPT-2 and that of BART-ELF22 when a randomly selected strategy was given.

\begin{figure}[!t]
\centering
\includegraphics[scale=0.305]{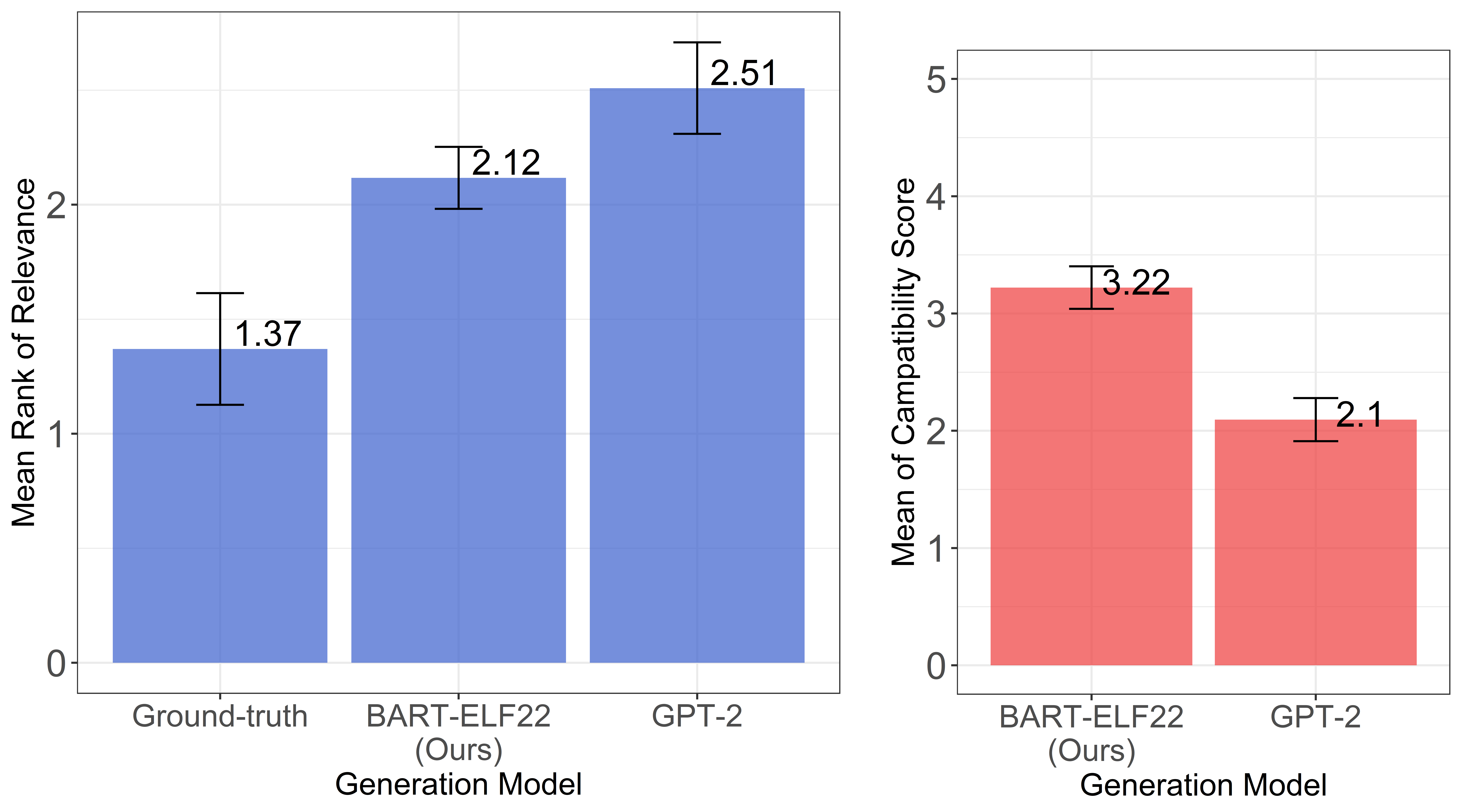}
\caption{Human evaluation results on response generation task C. In relevance (left), lower values indicate better performance, where the range is from \#1 to \#3. In compatibility (right), higher values indicate better performance, where the range is from 1 to 5.}
\label{fig:human_evaluation}
\end{figure}

Figure~\ref{fig:human_evaluation} shows the results of human evaluation. We confirm that BART-ELF22 (Ours) performed better than GPT-2 (baseline) in both relevance and compatibility.

For the relevance criterion, we compared three responses: ground-truth, a response by BART-ELF22 which was fine-tuned with our dataset, and GPT-2, which was not fine-tuned with our dataset. As shown in the left bar plot in Figure~\ref{fig:human_evaluation}, the mean values of ground-truth, BART-ELF22, and GPT-2 were 1.37, 2.12, and 2.51, respectively. According to Friedman test, there was a significant difference in mean rank between three responses ($\chi_{2}^{2}=10.00$, $p<.01$). For the post hoc comparison test, we used Wilcoxon Signed Ranks test. The result was in the order of ground-truth, BART–ELF22, and GPT-2, with significance established at $p<.05$. We speculate that fine-tuning with our dataset does improve the quality of the model to generate context-relevant sentences.

For the compatibility criterion, we compared two responses generated by BART-ELF22 and GPT-2. As shown in the right bar plot in Figure~\ref{fig:human_evaluation}, the mean values of BART-ELF22 and GPT-2 were 3.22 ($sd=0.18$) and 2.10 ($sd=0.18$), respectively. According to one-way repeated measure ANOVA test, there was a significant difference in the mean values between two responses ($F_{(1,4)} =41.8,  p<.01$). We see that fine-tuning with our dataset significantly improves the performance of strategy-controlled sentence generation.

\subsection{Case Study}

\begin{table}[t!]
\centering
\begin{tabular}{cc}
\toprule
\bf    & \bf Text \\
\midrule
\multicolumn{1}{l}{Title}  & \multicolumn{1}{l}{Smoking kill gain too much?}  \\
\hline
\multicolumn{1}{l}{Body}  & \multicolumn{1}{l}{\makecell[l]{I have been working out like crazy \\
and have been seeing good results\\
but I have seen that smoking can \\
kill testosterone cell. I dont smoke\\
that much like 3 cigerattes a day \\
will it affect my gains too much?}}   \\
\hline
\multicolumn{1}{l}{Comment}  & \multicolumn{1}{l}{\makecell[l]{
Everything is good with \\
moderation. Don’t go crazy and \\
everything should be fine.You \\
just...wont gain as much as you \\
normally would if you didn’t \\
smoke.
}} \\
\hline
\multicolumn{1}{l}{RL}  & \multicolumn{1}{l}{\makecell[l]{
in an expose way
}}   \\
\midrule
\multicolumn{1}{c}{\bf Model}        & \multicolumn{1}{c}{\bf Text} \\ 
\midrule
\multicolumn{1}{l}{GT}        & \multicolumn{1}{l}{\makecell[l]{
LOL smoking is not good with \\
moderation.
}} \\ 
\hline
\multicolumn{1}{l}{GPT-2}        & \multicolumn{1}{l}{\makecell[l]{
What is wrong with you? I don't \\
know why you are getting \\
downvoted lol
}} \\ 
\hline
\multicolumn{1}{l}{BART-ELF22}        & \multicolumn{1}{l}{\makecell[l]{
Are you talking about the same \\
thing?
}} \\ 
\hline
\multicolumn{1}{l}{\makecell[l]{BART-ELF22 \\ (engage)}}        & \multicolumn{1}{l}{\makecell[l]{
Well yes and no. I am not saying I \\
don’t see any results in \\
moderation, but I am saying this is \\
a general trend with smokers. If I \\
smoke that much that will not \\
affect my mental health.
}} \\ 
\hline
\multicolumn{1}{l}{\makecell[l]{BART-ELF22 \\ (challenge)}}    & \multicolumn{1}{l}{\makecell[l]{
Tell that to alcoholics. We all \\
know what we're getting into here.
}} \\ 
\bottomrule
\end{tabular}
\caption{Qualitative examples of pairs of troll comment and generated counter-responses by models. GT: Ground-truth; RL:
Response strategy label}
\label{tab:discussion_example}
\end{table}

Table~\ref{tab:discussion_example} shows examples of five corresponding sentences generated by the model when given contextual information and a response strategy labeled in our dataset. 
In Table~\ref{tab:discussion_example}, the three counter responses above are generated by the ground-truth reference, GPT-2, and BART-ELF22 models, respectively. The two counter responses below are generated by BART-ELF22 with the other strategies.

We find that the counter sentences generated by GPT-2 contain frequently repetitive words and composed of fewer informative tokens regardless of strategies. In BART-ELF22, as in GPT-2, there were some cases where the variation was small even if the given strategy was changed, but the frequency was relatively small compared to GPT-2. We observe that BART-ELF22 generates data by varying counter responses according to the specified strategy. This result suggests that the way the BART-ELF22 appends the strategy phrase-wise to the context does work. We see that the proposed dataset supports improvement on controlled response generation according to the response strategy.

\section{\label{sec:conclusions}Conclusions}
We presented a dataset in which types of trolls and response strategies are annotated against troll comments with the help of rich contextual information. We verified the effectiveness of our dataset from the experimental results, which showed that fine-tuning on our dataset improved the performances of all the tasks as well as the quality of counter responses. We also see that the fine-tuned text generation model generated data by varying counter responses according to the specified strategy, even when the specified strategy was not in our dataset. In future work, we will explore the shared features that are potentially obtained from counter speech datasets so as to improve the generation model. With this dataset, we hope to support the community in enabling healthy and open communication. The ELF22 dataset is publicly available at \href{https://github.com/huijelee/ELF22}{https://github.com/huijelee/ELF22}.

\section{Acknowledgements}

This work was supported by 
the National Research Foundation of Korea (NRF) (No. 2020R1A2C1010759, A system for offensive language detection and automatic feedback with correction) grant funded by the Korea government (MSIT).

\begin{table*}[t!]
\centering
\begin{tabular}{l|l|l|l|ll}
\hline
\multicolumn{1}{c|}{\bf Title} & 
\multicolumn{1}{c|}{\bf Post} & 
\multicolumn{1}{c|}{\bf Troll comment} & 
\multicolumn{1}{c|}{\bf Response} & 
\multicolumn{1}{c}{\bf TL} & 
\multicolumn{1}{c}{\bf RL} 
\\ \hline
\makecell[l]{
I’m so sorry guys,\\
but I can’t HODL\\
anymore
}    &
\makecell[l]{
Had to sell my DOGE,\\
lost my wife yesterday\\
to a heart attack, and\\
I’m going to need the\\
money to handle the\\
affairs that normally\\
come from the death of\\
a spouse. No, I’m not\\
ok, pretty far from it.\\
Sorry shibes.
}   & 
\makecell[l]{
Is that the rule : Rule\\
no 1 never invest\\
money you dont have?\\
Sorry not trolling
} & 
\makecell[l]{
I started with \$20, and\\
her death was very\\
sudden. Two incomes\\
down to one is harsh,\\
and I needed to cash\\
out.
}       
& 2      & 1     \\ \hline
\makecell[l]{
I have Xbox game\\
pass and I’m still\\
boredh
}    & 
\makecell[l]{
Any recommendations
}   & 
\makecell[l]{
You are what doctors\\
call ``fucking lazy"
} & 
\makecell[l]{
Thank you for your\\
input
}       
& 2      & 2      \\ \hline
\makecell[l]{
Women can't have\\
anything nice on\\
Reddit how come\\
when a guy does\\
literally the bare\\
minimum for a\\
female individual\\
they’re called a\\
“simp”
}    & 
\makecell[l]{
I post in aother sub\\
about how Deanna\\
Price became the\\
second woman ever\\
too throw over 80\\
meters in the hammer\\
throw. There are so\\
many comments about\\
her gender and\\
appearance, and all I\\
was trying to do was\\
celebrate this woman's\\
hard earned athletic\\
achievement. Two\\
weeks ago, someone\\
else posted in the\\
same sub about Ryan\\
Crouser setting a new\\
shot put word record,\\
and no one said shit\\
about his appearance\\
I hate it here sometimes
}   & 
\makecell[l]{
What? Get the fuck out\\
of here. I literally just\\
looked at your post\\
that you are talking\\
about and like 95\% of\\
all the comments are\\
positive.
} & 
\makecell[l]{
I never said they were\\
all negative, there's a\\
lot of nice comments,\\
but there are a lot of\\
really bad ones too.
}       
& 1      & 3      \\ \hline
\makecell[l]{
HS cheerleader\\
carwash\\
fundraisers\\ 
are creepy
}    & 
\makecell[l]{
I had to explain to my\\
16 year old daughter,\\
why her in a bikini\\
washing cars for a\\
public fundraiser was\\
not gonna be in our\\
plans. Come to find\\
out today after school\\
my daughter was glad\\
she didn’t go as in the\\
police were called on\\
someone that was\\
stalking the girls from\\
across the street and\\
making gestures for\\
about 40 minutes. The\\
whole concept is\\
weird why even\\
expose the girls like\\
that in the first place
}   & 
\makecell[l]{
It’s fine. What’s with\\
this recent trend of\\
making everything into\\
the worst possible\\
conclusion it could\\
possibly be? It’s just a\\
bunch of dads\\
checking out the\\
young budding\\
daughters of the other\\
dads in the town. You\\
know just like “oh hey\\
man, your daughter\\
has grown up so well,\\
she’s almost a woman\\
now, has she had her\\
first blood?” just dad\\
stuff.
} & 
\makecell[l]{
I hope you never have\\
kids, you fucking\\
weirdo. If someone\\
ever said that to me I\\
would punch their\\
lights out. If you have\\
a daughter I dare you\\
to show this comment\\
to her or try to explain\\
your thinking behind\\
it, and watch them\\
physically gag at your\\
creepy ass
}       
& 1      & 4      \\ \hline
\end{tabular}
\caption{Examples of collected Reddit posts. The numbers 1 and 2 of the TL column indicate overt and covert trolls, respectively. Response strategies from 1 to 4 of RL indicate \textit{engage},  \textit{ignore}, \textit{expose}, and \textit{challenge}, respectively.}
\label{tab:morepostexamples}
\end{table*}

\begin{table*}[t!]
\centering
\begin{tabular}{l|l|l|l|ll}
\hline
\multicolumn{1}{c|}{\bf Title} & 
\multicolumn{1}{c|}{\bf Post} & 
\multicolumn{1}{c|}{\bf Troll comment} & 
\multicolumn{1}{c|}{\bf Response} & 
\multicolumn{1}{c}{\bf TL} & 
\multicolumn{1}{c}{\bf RL} 
\\ \hline
\makecell[l]{
i am now actually\\ 
afraid of solo\\
queuing
}    & 
\makecell[l]{
valorant solo q is\\
actually scary to me\\
now. i've gone up\\
against way too many\\
smurfs and dealt with\\
so much toxicity its\\
actually scary to me\\
now. ranked is\\
something i actually\\
want to grind, but all\\
the problems with the\\
game is actually\\
worrying me. does\\
anyone else have this\\
problem or am i just\\
being fucking stupid?
}   & 
\makecell[l]{
Man you all sounds\\
so soft lmao
} & 
\makecell[l]{
Imagine a world\\
where gaming is only\\
filled with little toxic\\
shts who think it's\\
cool to just shoot out\\
insults just because\\
it's the internet.\\
Making the community\\
better only benefits\\
us. This isn't the past\\
anymore.
}       
& 2      & 5      \\ \hline
\makecell[l]{
How do I play\\
against mages?
}    & 
\makecell[l]{
I always feel like I am\\
being outranged and\\
can never get any\\
significant damage\\
down on them except\\
for W1. Furthermore, I\\
am always pushing\\
the lane with W, but I\\
am not sure how to\\
trade without it. How\\
should I be looking to\\
space against mages?
}   & 
\makecell[l]{
You don't. Just dodge\\
the game.
} & 
\makecell[l]{
dodge every mage im\\
the game ?XDDD
}       
& 2      & 6      \\ \hline
\makecell[l]{
I hate fireworks
}    & 
\makecell[l]{
Why are they even\\
legal? It scares the\\
shit out our pets and\\
our neighbors. It\\
leaves all kinds of\\
crap on the ground\\
and smoke is horrid.\\
How could they\\
legalize it???
}   & 
\makecell[l]{
JFC a bunch of\\
snowflakes here on\\
r/Ohio. Go live in\\
your basement by all\\
means. Especially if\\
you can’t comprehend\\
basic English while\\
living in Ohio. Your\\
grammar and spelling\\
sucks.
} & 
\makecell[l]{
I see you sub to\\
r/republican,\\
r/conservative,\\
as well as\\
r/AskAnAmerican\\
Are you then a proud\\
patriot?
}       
& 1      & 7      \\ \hline
\end{tabular}
\caption{Examples of collected Reddit posts. The numbers 1 and 2 of TL indicate overt and covert trolls, respectively. Response strategies from 5 to 7 of RL indicate \textit{critique}, \textit{mock}, and \textit{reciprocate}, respectively.}
\label{tab:morepostexamples2}
\end{table*}

\section{Bibliographical References}\label{reference}

\bibliographystyle{lrec2022-bib}
\bibliography{lrec2022}

\section*{Appendix: More Examples}

Tables~\ref{tab:morepostexamples} and~\ref{tab:morepostexamples2} show examples of collected Reddit posts, along with annotated strategies.

\end{document}